\documentclass{llncs}

\usepackage[pdftex]{graphicx}

\usepackage[T1]{fontenc}
\usepackage[utf8]{inputenc}

\begin{document} 

\title{Neighborhood selection and rules identification for cellular automata: a rough sets approach}
\titlerunning{Neighborhood selection and rules identification for cellular automata}
\toctitle{Neighborhood selection and rules identification for cellular automata: a rough sets approach}

\author{Bartłomiej Płaczek}
\institute{Institute of Computer Science, University of Silesia,\\
Będzińska 39, 41-200 Sosnowiec, Poland \\
\email{Placzek.Bartlomiej@gmail.com}}

\maketitle

\begin{abstract}
In this paper a method is proposed which uses data mining techniques based on rough sets theory to select neighborhood and determine update rule for cellular automata (CA). According to the proposed approach, neighborhood is detected by reducts calculations and a rule-learning algorithm is applied to induce a set of decision rules that define the evolution of CA. Experiments were performed with use of synthetic as well as real-world data sets. The results show that the introduced method allows identification of both deterministic and probabilistic CA-based models of real-world phenomena. \footnote{Preprint of: Płaczek B.: Neighborhood selection and rules identification for cellular automata: a rough sets approach. Lecture Notes in Computer Science, vol. 8385, pp. 721-730 (2014). The final publication is available at www.springerlink.com}
\end{abstract}
\section{Introduction}
Cellular automata (CA) have found many applications in the field of complex systems modeling. There is a number of works devoted to CA models for simulation of real-world phenomena like pedestrian dynamics \cite{bibbp15}, traffic flow \cite{bibbp10}, urban growth \cite{bibbp7}, etc. In most cases, the update rule and neighborhood for CA are determined by human experts that have knowledge of the modeled phenomenon. Automatic identification of CA-based models remains an open research issue. 

Several attempts have been made in the literature to develop algorithms for CA identification. However, most of the previous research did not investigate the use of real-world data sets as an input. The available algorithms were designed and tested mainly against synthetic data obtained from CA evolution. The use of real-world data for identification of CA models poses additional challenges due to inherent complexity of modeled phenomena and errors that are made during data acquisition.

This paper discusses the possibility of using data mining techniques based on the rough sets theory \cite{bibbp3} to select neighborhood and determine rules for CA models. According to the proposed approach, input data describing observed states of cells are represented in form of a decision table. Neighborhood is detected by using algorithms for reducts calculation. Cells that do not belong to the neighborhood are removed from the decision table. After that, a rule-learning method is applied to induce a set of decision rules that define the evolution of CA. This approach was tested in identification of both deterministic and probabilistic CA models for synthetic as well as real-world data sets.
\section{Related works}
 Most of the works related to the CA identification problem use genetic algorithms as a tool to extract update rule and neighborhood from spatio-temporal patterns produced in CA evolution \cite{bibbp4,bibbp8,bibbp16}. In \cite{bibbp11} a genetic algorithm was employed to learn probabilistic rules directly from experimental data for two-dimensional binary-state CA. Several methods were proposed that use genetic algorithms to learn CA rules and neighborhood size for image processing tasks \cite{bibbp5,bibbp6}.

Although application of genetic algorithms was a dominant approach to CA identification, some non-genetic techniques are also available. Adamatzky \cite{bibbp2} proposed several approaches to extracting rules for different classes of CA. Straatman et al. \cite{bibbp13} developed a form of hill climbing and backtracking to identify CA rules. In \cite{bibbp12} a~deterministic sequential floating forward search method was used to select rules of CA. Another approach is based on parameter estimation methods from the field of system identification \cite{bibbp17}. A framework for solving the identification problems for both deterministic and probabilistic CA was presented in \cite{bibbp14}.

 Main drawbacks of the existing CA identification methods are related to the fact that they were either designed for a particular class of CA, or their experimental evaluation was limited to the case of synthetic data. The introductory research results presented in this paper shows that the rough sets approach may be effectively used to develop a universal method for identification of CA models that mimic real-world phenomena.
\section{Basic concepts}
Formally, a cellular automaton can be defined as a triple $(V, N, \delta)$, where $V$ is a non-empty set of cell states, $N$ is the neighbourhood, and $\delta$ is the update rule. Arguments of $\delta$ are the current states of cells in the neighbourhood, while the value of $\delta$ indicates the state of a central cell at the next time step.

The problem of CA identification involves finding both the cells neighborhood and the update rule on the basis of a training data set, which includes observed states of the cells. In order to use the rough sets approach for solving this problem, the training set of data has to be represented in the form of a decision table $I=(U,C)$, where $U$ is a non-empty set of observations and $C$ is a set of cell states: $C=\{c_{\alpha}(t),\dots,c_{i}(t),\dots,c_{\omega}(t),c_{i}(t+1)\}$.

State of $j$-th cell at time step $t$ is denoted by $c_{j}(t)$. Index $i$ indicates the central cell, for which neighborhood and update rule have to be found. Thus, the cell state $c_{i}(t+1)$ is used as the decision attribute. The remaining cell states $c_{\alpha}(t),\dots,c_{\omega}(t)$ are condition attributes. The candidate neighbors $\alpha,\dots ,\omega$ are the cells for which distance to the central cell $i$ is lower than a threshold value. The threshold is determined experimentally. Above definition can be easily extended to multidimensional CA by adding the necessary indexes. 

Neighborhood for the $i$-th cell can be selected by calculating reducts of the above-defined decision table. A reduct is a subset of the condition attributes, which is sufficient to determine the decision attributes \cite{bibbp3}. Taking into account the decision table discussed above, reduct should be defined as a subset of cell states $R \subseteq c_{\alpha}(t),\dots,c_{\omega}(t)$, which preserves discernibility of the observations with regard to the decision $c_{i}(t+1)$, and none of its proper subsets has this ability. Observations are discernible if they differ in at least one condition attribute (cell state). Each two observations that have different decisions $c_{i}(t+1)$ and are discernible by the full set of cell states $c_{\alpha}(t),\dots ,c_{\omega}(t)$ are also discernible by the reduct $R$. 

The neighborhood $N$ of a cell is determined as a set of cells, whose states belongs to the reduct $R: N=\{j: c_{j}(t) \in R\}$, where $j$ is a cell index. There may exist multiple reducts for one decision table. Selection of the neighborhood is made with regard to the shortest reduct, because size of the neighborhood has to be minimized. If there are several minimal reducts then the one is selected which has the lowest average distance between neighbors and the central cell.
When the reduct $R$ is found, the condition attributes that do not belong to this reduct are excluded from the decision table. Thus, the modified decision table $I'=(U,C')$ has the following set of attributes: $C'=R \cup \{c_{i}(t+1)\}$, where $c_{i}(t+1)$ remains the decision attribute.

Update rule of a cellular automaton is identified as a set of decision rules by taking into account the information from the modified decision table $I'$. A particular decision rule $r$ has the following form:

\begin{equation}
(c_{x}(t)=v_{x}) \wedge \dots \wedge (c_{y}(t)=v_{y}) \Rightarrow c_{i}(t+1)=v_{i}
\end{equation}
where: $\{c_{x}(t),\dots,c_{y}(t)\} \subseteq R$, $v_{j}\in V$, and $V$ denotes the set of allowable cell states.

Two characteristics of decision rules are useful for the proposed method: support and match \cite{bibbp3}. Support of rule $r$, denoted by $\textit{SUPP}_{I'}(r)$, is equal to the number of observations from $I'$ for which rule $r$ applies correctly, i.e., premise of the rule is satisfied and the decision given by rule is consistent with the one in decision table. $\textit{MATCH}_{I'}(r)$ is the number of observations in $I'$ for which the rule $r$ applies, i.e., premise of the rule is satisfied. Based on these two characteristics, a certainty factor is defined for the decision rule $r$:

\begin{equation}
\textit{CER}_{I'}(r)=\textit{SUPP}_{I'}(r)/\textit{MATCH}_{I'}(r).
\end{equation}

The certainty factor may be interpreted as a conditional probability that decision of rule $r$ is consistent with an observation in the decision table $I'$, given that premise of the rule is satisfied.

Fig. 1 summarizes the main operations that are necessary to identify a cellular automaton by using the rough sets approach. In this study, reducts and decision rules are calculated using algorithms implemented in the RSES software \cite{bibbp3}. Three algorithms of reducts calculation were examined: exhaustive, genetic, and dynamic reduct algorithm. Moreover, the experiments involved application of three algorithms that enable induction of decision rules: exhaustive, genetic, and LEM2. 
\begin{figure}
\centering
\includegraphics[width=10cm]{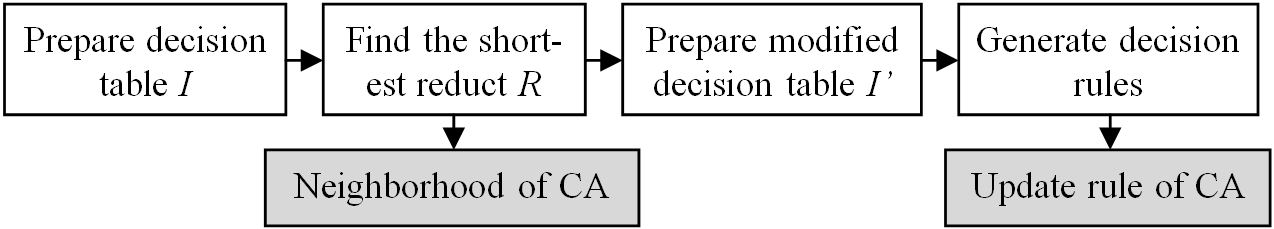}
\caption{Rough sets based procedure of CA identification}
\end{figure}
\section{Identification of deterministic cellular automata}
In this section, the proposed approach is applied for identification of three deterministic CA \cite{bibbp9}: elementary cellular automaton with Wolfram's rule 184 (ECA-184), deterministic version of Nagel-Schreckenberg cellular automaton (NaSh-D), and the Conway's game of life cellular automaton (Life).

ECA-184 is a one-dimensional cellular automaton with binary cell states and neighborhood of three cells wide. Original definition of the ECA-184 update rule is presented in Fig. 2. The upper row in this figure illustrates all possible states of a central cell and its neighborhood. Lower row shows states of the central cell after update -- in the next time step of the CA evolution.
\begin{figure}
\centering
\includegraphics[width=10cm]{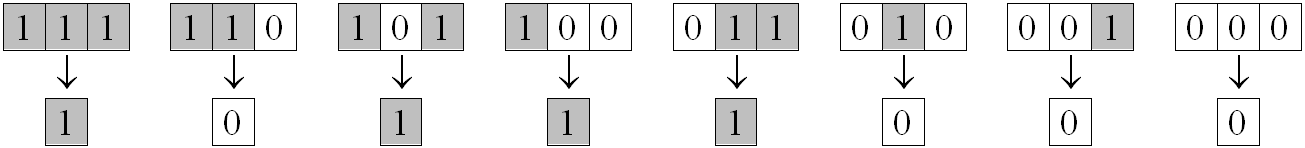}
\caption{Update rule of ECA-184}
\end{figure}

Identification of ECA-184 was performed using a training data set of 500 observations. Each observation in the decision table $I$ has covered a group of 21 cells (candidate neighbors) and the 11-th cell was considered as the central one. The neighborhood of ECA-184 was correctly recognized by each of the reducts calculation algorithms (exhaustive, genetic, and dynamic). The shortest reduct was determined as $R=\{c_{i-1}(t),c_{i}(t),c_{i+1}(t)\}$, thus the modified decision table $I'$ had four attributes: $C'=\{c_{i-1}(t),c_{i}(t),c_{i+1}(t),c_{i}(t+1)\}$. Tab. 1 presents the decision rules that were calculated from table $I'$ by using the LEM2 algorithm. Symbol $\emptyset$ indicates that a given cell state does not occur in a particular decision rule, e.g., the rule no. 3 from Tab. 1 should be interpreted as: $(c_{i-1}(t)=1)\wedge(c_{i}(t)=0)\Rightarrow c_{i}(t+1)=1$. The set of decision rules in Tab. 1 is consistent with the original update rule of ECA-184 (Fig. 2). The remaining algorithms of rule induction (exhaustive and genetic) have also generated valid sets of decision rules; however their size was larger (6 rules).  
\begin{table}
\caption{Decision rules generated for ECA-184}
\begin{center}
\setlength{\tabcolsep}{6pt}
\begin{tabular}{|c|c|c|c|c|c|} \hline 
Rule no. & 1 & 2 & 3 & 4 & 5 \\ \hline 
$c_{i-1}(t)$ & 0 & 0 & 1 & $\emptyset$ & 1 \\ \hline 
$c_{i}(t)$ & $\emptyset$ & 0 & 0 & 1 & 1 \\ \hline 
$c_{i+1}(t)$ & 0 & $\emptyset$ & $\emptyset$ & 1 & 0 \\ \hline 
$c_{i}(t+1)$ & 0 & 0 & 1 & 1 & 0 \\ \hline  
\end{tabular}
\end{center}
\end{table}

The second CA identification example concerns a deterministic version of the Nagel-Schreckenberg model for road traffic simulation (NaSch-D) \cite{bibbp9}. Update rule of NaSch-D consists of two steps: (I) acceleration and braking of vehicle, (II) vehicle movement. In step I velocity $v_{k}(t)$ for each vehicle $(k)$ is calculated in cells per time step: $v_{k}(t)\leftarrow \min\{v_{k}(t-1)+ 1,g_{k}(t-1),v_{\max}\}$, where $g_{k}$ is the number of empty cells in front of vehicle $k$,  $v_{\max}$ denotes maximal velocity of vehicles. Step II simulates movement of the vehicles -- index of a cell occupied by vehicle \textit{k} at time step \textit{t}, denoted by $x_{k}(t)$, is determined using the formula $x_{k}(t)\leftarrow x_{k}(t-1)+v_{k}(t)$.

In this study, the parameter $v_{max}$ was set to 2 cells per time step. Thus, the set of cell states includes 4 values: $c_{i}(t)=-1$ denotes empty cell, and $c_{i}(t)=0,\ldots,2$ indicates velocity of the vehicle that occupies the $i$-th cell. The training data set was prepared in the same way as for the previous example. Decision table $I$ contained 2000 observations. The shortest reduct $R=\{c_{i-2}(t),\dots,c_{i+2}(t)\}$ was uniquely determined by all the examined algorithms. Tab. 2 shows the set of 16 decision rules generated for NaSch-D by using LEM2 algorithm. In case of exhaustive as well as genetic algorithm the number of obtained decision rules was 37.

\begin{table}
\caption{Decision rules generated for NaSch-D}
\begin{center}
\setlength{\tabcolsep}{5pt}
\begin{tabular}{|c|c|c|c|c|c|c|c|c|c|c|c|c|c|c|c|c|} \hline 
Rule no. & 1 & 2 & 3 & 4 & 5 & 6 & 7 & 8 & 9 & 10 & 11 & 12 & 13 & 14 & 15 & 16\\ \hline 
$c_{i-2}(t)$ & -1 & -1 & $\emptyset$ & $\emptyset$ & 0 & $\emptyset$ & 2 & 1 & $\emptyset$ & 2 & $\emptyset$ & 2 & 2 & 2 & 2 & 2 \\ \hline  
$c_{i-1}(t)$ & $\emptyset$ & -1 & $\emptyset$ & $\emptyset$ & $\emptyset$ & 1 & -1 & -1 & 1 & -1 & 1 & -1 & -1 & -1 & -1 & -1 \\ \hline  
$c_{i}(t)$ & $\emptyset$ & -1 & 0 & 0 & $\emptyset$ & -1 & -1 & $\emptyset$ & -1 & -1 & -1 & -1 & -1 & -1 & -1 & -1 \\ \hline
$c_{i+1}(t)$ & -1 & $\emptyset$ & 0 & $\emptyset$ & -1 & 0 & -1 & -1 & 2 & 0 & 1 & 2 & -1 & -1 & 1 & -1 \\ \hline
$c_{i+2}(t)$ & $\emptyset$ & $\emptyset$ & $\emptyset$ & -1 & $\emptyset$ & $\emptyset$ & -1 & $\emptyset$ & -1 & $\emptyset$ & -1 & $\emptyset$ & 2 & 0 & $\emptyset$ & 1 \\ \hline
$c_{i}(t+1)$ & -1 & -1 & 0 & 1 & -1 & 0 & 2 & -1 & 2 & 0 & 1 & 2 & 2 & 1 & 1 & 2 \\ \hline
\end{tabular}
\end{center}
\end{table}

The set of decision rules in Tab. 2 has the smallest size. However, its application requires additional assumptions to resolve conflicts when various rules that match the observed state of the neighborhood suggest different states of the central cell. This issue was addressed by using priorities of the decision rules. In case of conflict, the rule with the highest priority is selected. The priority is determined taking into account support of the rule. A rule with a larger support has higher priority. In Tab.~2, the decision rules are sorted in descending order, according to their priorities. It was verified that for each possible state of the neighborhood, the above defined decision algorithm enables correct update of the central cell.

 Next, the proposed method was applied to identify the two dimensional CA called Life \cite{bibbp9}. In this example, the binary cell states $c_{i,j}(t)=0$ or $c_{i,j}(t)=0$ indicates that the cell $(i,j)$ is dead or alive respectively. Every cell interacts with its eight neighbors (Moore neighborhood). At each time step, the following updates occur: (1) Any live cell with fewer than two live neighbors dies. (2) Any live cell with two or three live neighbors lives on to the next generation. (3) Any live cell with more than three live neighbors dies. (4) Any dead cell with exactly three live neighbors becomes a live cell.

 Observations collected in a decision table for identification of the Life CA have included states of 25 cells from the Moore neighborhood of radius 2. The actual neighborhood that includes 9 cells was correctly recognized by the reducts calculation algorithms. However, taking into account the states of all 9 cells from the neighborhood, a large set of decision rules was obtained (154-242 rules, depending on the algorithm). In order to decrease the number of decision rules, the amount of live neighbors $L$ was added as an attribute of the decision table. After this modification, the reducts calculation algorithms were executed again, and the resulting shortest reduct has included only two attributes: $R=\{c_{i,j}(t),L\}$. Thus, a decision table $I'$ with attributes $C'=\{c_{i,j}(t),L,c_{i,j}(t+1)\}$ was used to generate decision rules. The set of rules in Tab. 3 was generated by the exhaustive rule induction algorithm. These decision rules are consistent with the update rule of Life, which was described above. In case of LEM2 algorithm, the rule set had 17 elements. The genetic algorithm has provided an incomplete set of decision rules that fail to describe the evolution of Life.
 
 \begin{table}
 \caption{Decision rules generated for Life}
 \begin{center}
 \setlength{\tabcolsep}{6pt}
 \begin{tabular}{|c|c|c|c|c|c|c|c|c|c|c|} \hline 
 Rule no. & 1 & 2 & 3 & 4 & 5 & 6 & 7 & 8 & 9 & 10 \\ \hline 
 $c_{i,j}(t)$ & $\emptyset$  & $\emptyset$  & 0 & 1 & $\emptyset$ & $\emptyset$ & $\emptyset$ & $\emptyset$ & $\emptyset$ & $\emptyset$  \\ \hline 
 $L$ & 0 & 1 & 2 & 2 & 3 & 4 & 5 & 6 & 7 & 8 \\ \hline 
 $c_{i,j}(t+1)$ & 0 & 0 & 0 & 1 & 1 & 0 & 0 & 0 & 0 & 0\\ \hline 
 \end{tabular}
 \end{center}
 \end{table}
\section{Identification of probabilistic cellular automata}
 In case of probabilistic CA, the identification task is more complicated because the state of neighborhood does not uniquely determine the state of central cell. Therefore, some of the decision rules that describe evolution of a probabilistic CA are uncertain. It means that there exist decision rules that have identical premises and give different decisions. According to the proposed approach, such rules are merged into one rule whose decision is defined using a set of pairs of vales $v_{z}$ and certainty factors $\textit{CER}_{I'}(r_{z})$:
\begin{equation}
 (c_{x}(t)=v_{x}) \wedge \dots \wedge (c_{y}(t)=v_{y}) \Rightarrow c_{i}(t+1)=\{v_{z}/\textit{CER}_{I'}(r_{z})\},
\end{equation}

\noindent where $r_{z}$ indicates the decision rule $(c_{x}(t)=v_{x}) \wedge \dots \wedge (c_{y}(t)=v_{y}) \Rightarrow c_{i}(t+1)=v_{z}$. 

During update operation of a CA, if the premise $(c_{x}(t)=v_{x}) \wedge \dots \wedge (c_{y}(t)=v_{y})$ is satisfied then the state of a central cell $c_{i}(t+1)$ takes value $v_{z}$ with probability $\textit{CER}_{I'}(r_{z})$. This method utilizes the fact that certainty factor $\textit{CER}_{I'}(r_{z})$ may be interpreted as a conditional probability (see section 3). It should be also noted that for the rule defined by (3) the condition $\Sigma_{z}\textit{CER}_{I'}(r_{z})=1$ is always satisfied. 

The rough sets approach was applied to identification of the Nagel-Schrecken-berg probabilistic cellular automaton (NaSch) \cite{bibbp9}. Deterministic version of NaSch was described in section 4. For probabilistic NaSch, the first step of update rule is extended by step I-a, which includes randomization of the velocity according to formula $\xi(t)<p\Rightarrow v_{k}(t)\leftarrow \max\{0,v_{k}(t)-1\}$, where: $\xi\in[0,1)$ is a random number drawn from a uniform distribution, and $p\in[0,1]$ is a parameter called braking probability.

Traffic simulation was performed by using the NaSch model with parameters $v_{max}=1$ and $p=0.2$. The binary cell states were used in this example to distinguish empty and occupied cells. A decision table was prepared based on observations that were collected during the simulation. Reducts of the decision table were calculated to select the neighborhood. Due to the existence of conflicting observations, the genetic algorithm did not find any reducts and reducts calculated by both the exhaustive and the dynamic algorithm have included a number of cell states that in fact are not taken into account by the update rule of the analyzed CA. It was necessary to use a reduct shortening algorithm \cite{bibbp3} for correct determination of the neighborhood. After shortening operation, the reduct $R=\{c_{i-1}(t),c_{i}(t),c_{i+1}(t)\}$ was found, which is consistent with the actual neighborhood of cells in NaSch. The set of decision rules generated by LEM2 algorithm (Tab. 4) is equivalent to the update rule of NaSch. For remaining algorithms incomplete sets of the decision rules were obtained.

 \begin{table}
 \caption{Decision rules generated for NaSch ($v_{max}=1$)}
 \begin{center}
 \setlength{\tabcolsep}{4pt}
 \begin{tabular}{|c|c|c|c|c|c|c|c|} \hline 
 Rule no. & 1 & 2 & 3 & 4 & 5 & 6 & 7 \\ \hline 
 $c_{i-1}(t)$ & 0 & 0 & 0 & 1 & 1 & 1 & 1 \\ \hline
 $c_{i}(t)$ & 0 & 1 & 1 & 0 & 0 & 1 & 1 \\ \hline
 $c_{i+1}(t)$ & $\emptyset$ & 0 & 1 & 0 & 1 & 0 & 1 \\ \hline
 $c_{i}(t+1)$ & 0/1 & 0/0.8, 1/0.2 & 1/1 & 0/0.2, 1/0.8 & 0/0.2, 1/0.8 & 0/0.8, 1/0.2 & 1/1  \\ \hline
 \end{tabular}
 \end{center}
 \end{table}

The proposed approach was also verified in identification of NaSch model with parameter $v_{max}=2$, however the results are not presented here due to space limitations.

\section{Identification of cellular automata using real-world data}
In this section the rough sets approach is used to identify a CA on the basis of real-world observations regarding shock waves that emerge in highway traffic. The analyzed traffic data, describing vehicles trajectories in time space diagrams, were taken from \cite{bibbp18}. Training data set includes cell states that correspond to the positions of vehicles determined from the time space diagrams. It was assumed that one cell represents a 7 m segment of traffic lane. State of a cell is defined as a binary value: 0 denotes empty cell and 1 refers to an occupied cell. An example of the training data is presented in Fig. 3 a).

Neighborhood and update rule were determined using the method devised for probabilistic CA in previous section. The algorithm for finding dynamic reducts was applied and followed by the reduct shortening operation. The resulting reduct includes states of four cells: $R=\{c_{i-2}(t),$ $c_{i-1}(t),$ $c_{i}(t),$ $c_{i+1}(t)\}$. Decision rules were generated by using the LEM2 algorithm. The obtained rules (Tab. 5) do not guarantee that the number of occupied cells (vehicles) will be constant during the CA evolution. This fact leads to unrealistic traffic simulation. 

 \begin{table}
 \caption{Decision rules generated from real-world traffic data}
 \begin{center}
 \setlength{\tabcolsep}{6pt}
 \begin{tabular}{|c|c|c|c|c|c|c|c|c|} \hline 
 Rule no. & 1 & 2 & 3 & 4 & 5 & 6 & 7 & 8 \\ \hline 
  $c_{i-2}(t)$ & 0 & 0 & 0 & 0 & 0 & 0 & 0 & 0 \\ \hline
  $c_{i-1}(t)$ & 0 & 0 & 0 & 0 & 1 & 1 & 1 & 1 \\ \hline
  $c_{i}(t)$ & 0 & 0 & 1 & 1 & 0 & 0 & 1 & 1 \\ \hline
  $c_{i+1}(t)$ & 0 & 1 & 0 & 1 & 0 & 1 & 0 & 1 \\ \hline
  $c_{i}(t+1)$ & 0/1 & 0/1 & 0/1 & 1/1 & 0/0.35, & 0/0.2, & 0/0.5,& 1/1  \\ 
   & & & & & 1/0.65 & 1/0.8 & 1/0.5 & \\ \hline \hline
 Rule no. & 9 & 10 & 11 & 12 & 13 & 14 & 15 & 16 \\ \hline 
  $c_{i-2}(t)$ & 1 & 1 & 1 & 1 & 1 & 1 & 1 & 1 \\ \hline
  $c_{i-1}(t)$ & 0 & 0 & 0 & 0 & 1 & 1 & 1 & 1 \\ \hline
  $c_{i}(t)$ & 0 & 0 & 1 & 1 & 0 & 0 & 1 & 1 \\ \hline
  $c_{i+1}(t)$ & 0 & 1 & 0 & 1 & 0 & 1 & 0 & 1  \\ \hline
  $c_{i}(t+1)$ & 0/0.75, & 0/0.75, & 0/0.8, & 1/1 & 0/0.5, & 0/0.5, & 0/0.3, & 1/1 \\ 
   &  1/0.25 & 1/0.25 & 1/0.2 & & 1/0.5 & 1/0.5 & 1/0.7 & \\ \hline
 \end{tabular}
 \end{center}
 \end{table}

The unrealistic model behavior may occur because the decisions for uncertain rules are made randomly, without any coordination. E.g., the rules 7 and 13 in Tab. 5 apply to the same configuration of cells $(0,1,1,0,0)$. If for both rules the random decision is 1 then the resulting configuration $(0,1,1,1,0)$ includes three instead of two occupied cells. This problem was resolved by adding error detection rules to the CA update algorithm. If error is detected by rules (4) then the update of cells $(i-2,i-1,i)$ is repeated. 

\begin{equation}
\begin{array}{l}
(c_{i-2}(t)=1)\wedge(c_{i-1}(t)=0)\wedge(c_{i}(t)=0)\wedge \\ \wedge(c_{i-2}(t+1)+c_{i-1}(t+1)+c_{i}(t+1)\neq 1) \Rightarrow \mbox{error}=\mbox{true} \\ (c_{i-2}(t)=1)\wedge(c_{i-1}(t)=0)\wedge(c_{i}(t)=1)\wedge \\ \wedge(c_{i-2}(t+1)+c_{i-1}(t+1)\neq1) \Rightarrow \mbox{error}=\mbox{true}
\end{array}
\end{equation}

Above modifications allow the CA to perform realistic simulation of traffic flow. Fig. 3 presents time space diagrams for two shock waves. The shock wave a) was observed in real highway traffic and the shock wave b) was obtained from evolution of the identified CA. This example illustrates the possibility of using the rough sets approach for identification of CA-based models of real-world phenomena.

\begin{figure}
\centering
\includegraphics[width=10cm]{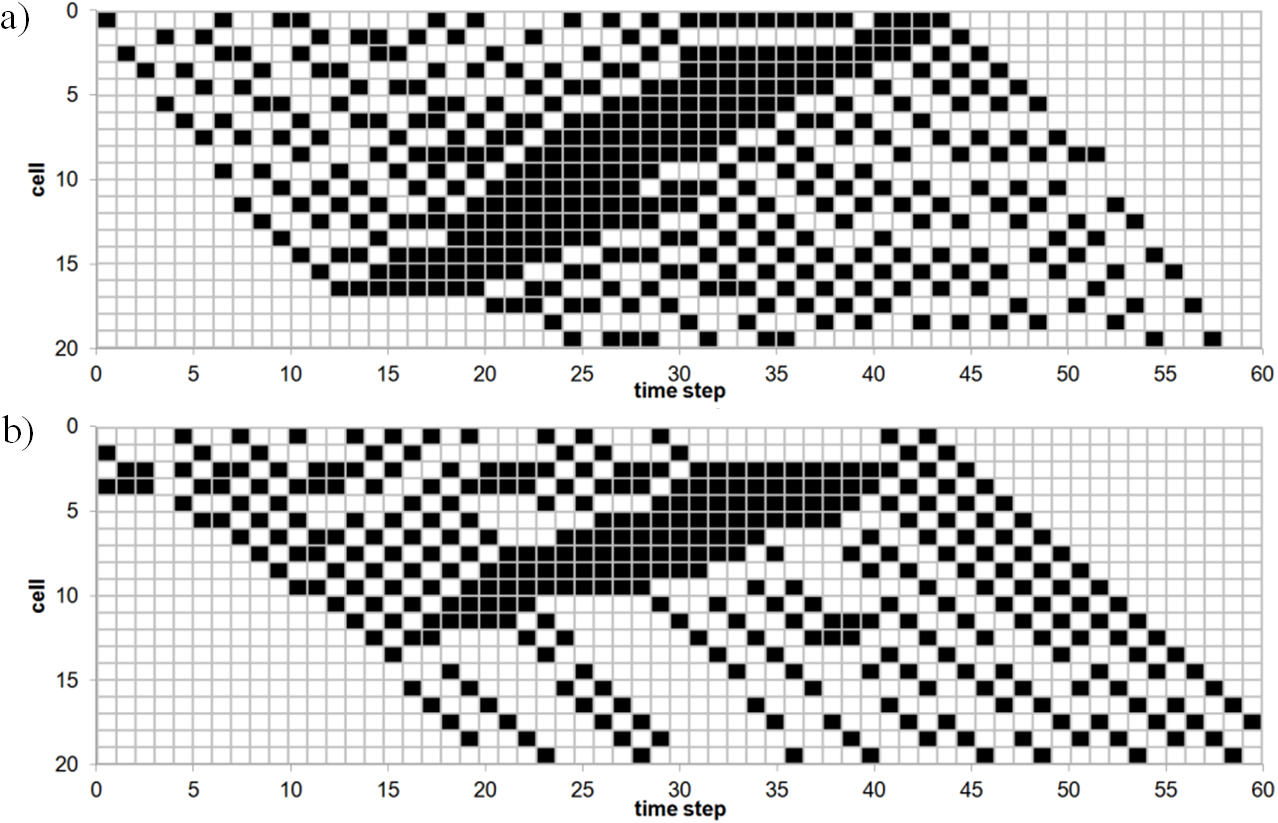}
\caption{Shock waves: a) real-world data, b) results of CA evolution}
\end{figure}

\section{Conclusions}
Results of the introductory research show that the data exploration techniques based on rough sets theory enable proper selection of neighborhood and update rule induction for different classes of CA. It was also demonstrated that the rough sets approach is suitable for CA models identification from real-world data sets. The best results were obtained for the proposed CA identification procedure when implemented by using the dynamic reducts algorithm for neighborhood selection and the LEM2 algorithm for decision rules induction. Nevertheless, in case of stochastic CA identification, some additional actions were necessary (reduct shortening, adding error detection rules). Verification of the usability of the proposed approach for a wider set of data as well as comparison with state-of-art methods based on genetic algorithms remain open issues for further research. Another interesting topic for future studies is to apply the proposed method in a data exploration system for CA-based image processing \cite{bibbp19,bibbp20}.

\end{document}